\documentclass[11pt]{article}

\usepackage[preprint]{acl}

\usepackage{times}
\usepackage{latexsym}

\usepackage[T1]{fontenc}

\usepackage[utf8]{inputenc}

\usepackage{microtype}

\usepackage{inconsolata}

\usepackage{graphicx}
\usepackage{amsmath}
\usepackage{amsfonts}
\usepackage{algorithm}
\usepackage{algpseudocode}
\usepackage{booktabs}
\usepackage{adjustbox}
\usepackage{multirow}
\usepackage{colortbl}
\usepackage{arydshln}
\usepackage{float}

\usepackage{caption}
\usepackage{tcolorbox}
\tcbuselibrary{breakable}

\usepackage{array}
\usepackage[table]{xcolor}
\usepackage{arydshln}

\title{ProQuant: Preserving Expert Quality over Quantity in Ultra-Low-Bit Quantization of Mixture-of-Experts Models}

\author{
 \textbf{Jiaqi Chen\textsuperscript{1}}\thanks{\; Equal Contribution.},
 \textbf{Xianglong Yan\textsuperscript{1}}\footnotemark[1],
 \textbf{Chengzhu Bao\textsuperscript{1}},
 \textbf{Xie Chen\textsuperscript{1}},
 \textbf{Yulun Zhang\textsuperscript{1}}\thanks{\; Corresponding author: Yulun Zhang.}
\\
 \textsuperscript{1}Shanghai Jiao Tong University
\\
 \small{
   \textbf{Correspondence:} \href{mailto:email@domain}{yulun100@gmail.com}
 }
}

\begin{document}
\maketitle


\begin{abstract}
\vspace{-1mm}
Mixture-of-Experts (MoE) large language models reduce computation through sparse activation but still incur substantial storage costs. Mixed-precision post-training quantization (PTQ) reduces these costs by quantizing different model components at different bit widths, but existing methods often struggle to preserve generation quality under tight bit budgets. We identify two key challenges: (1) nonlinear quantization errors at extremely low precision are underestimated when cross-layer propagation is ignored; (2) direct router fine-tuning may cause routing drift, and calibration gains may not generalize to downstream tasks. To address these issues, we propose ProQuant, a weight-only PTQ framework for MoE models. Prune-or-Prioritize Mixed-Precision Allocation (PoP) jointly optimizes expert pruning and bit-width assignment. It allows higher-precision experts to substitute for severely degraded ones while improving bit-budget utilization. Routing-Consistent RMSNorm Finetune (RoCo) adapts expert input representations through mathematically equivalent transformations. It compensates for quantization errors to reduce routing drift. Extensive experiments demonstrate that ProQuant outperforms existing methods across diverse generative tasks. Code is available at \url{https://github.com/ThrowBug/ProQuant.git}.

\end{abstract}

\setlength{\abovedisplayskip}{2pt}
\setlength{\belowdisplayskip}{2pt}

\begin{figure}[t]
    \centering
    \includegraphics[width=\linewidth]{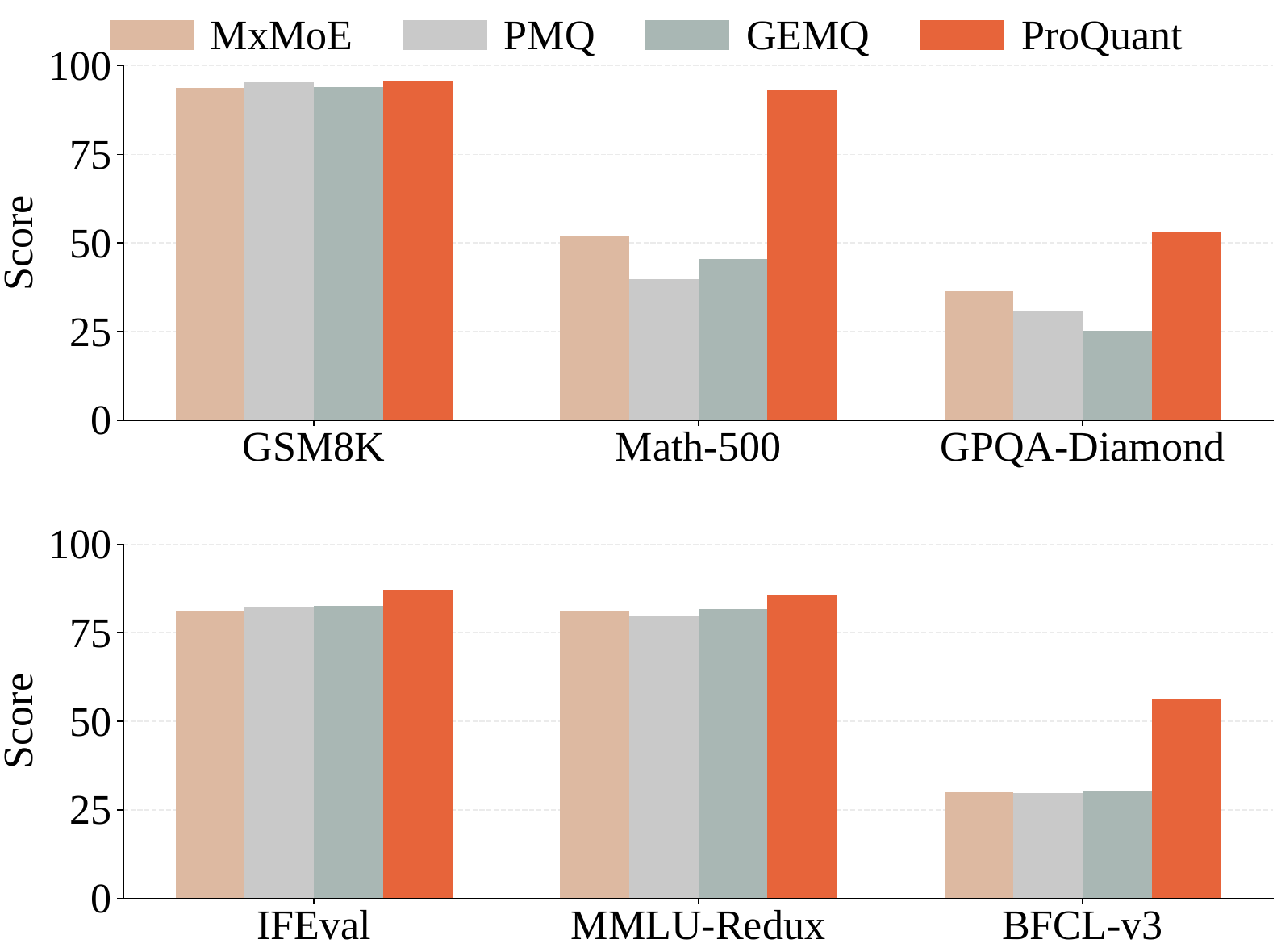}
    \vspace{-6mm}
    \caption{Performance comparison of MoE mixed-precision quantization methods on Qwen3-30B-A3B-Instruct-2507 under 2.5-bit average expert quantization.}
    \label{fig:performance}
    \vspace{-6mm}
\end{figure}


\vspace{-3mm}
\section{Introduction}
\vspace{-1mm}
Large language models (LLMs)~\citep{team2026kimi,zeng2026glm,chen2026minimax,blakeman2026nemotron,agarwal2025gpt} have demonstrated remarkable capabilities in natural language understanding and complex reasoning, while scaling model capacity remains an important path toward further performance improvements. Mixture-of-Experts (MoE) architectures~\citep{yang2025qwen3,liu2024deepseek,xu2026deepseek} increase model capacity by activating only a small subset of experts for each token, maintaining relatively low active parameters and computation. However, sparse activation does not reduce the need to store the full set of expert weights. As MoE models continue to scale, the resulting storage, memory, and weight-movement costs increasingly become major bottlenecks for efficient deployment.

To address these deployment challenges, weight-only post-training quantization (PTQ)~\citep{frantar2022gptq,lin2024awq,dettmers2023spqr,kim2023squeezellm,pmlr-v235-tseng24a} provides an efficient solution for compressing large-scale MoE models. Since autoregressive decoding is typically memory-bandwidth bound, reducing weight precision lowers both storage requirements and weight transfer costs during inference. Compared with quantization-aware training (QAT)~\citep{liu2024llm,chen2025efficientqat,du2024bitdistiller,liu2026paretoq,panferov2025quest}, PTQ requires no large-scale retraining and only a small amount of calibration data, making it particularly suitable for efficient MoE deployment.

\begin{figure}[t]
    \centering
    \includegraphics[width=\linewidth]{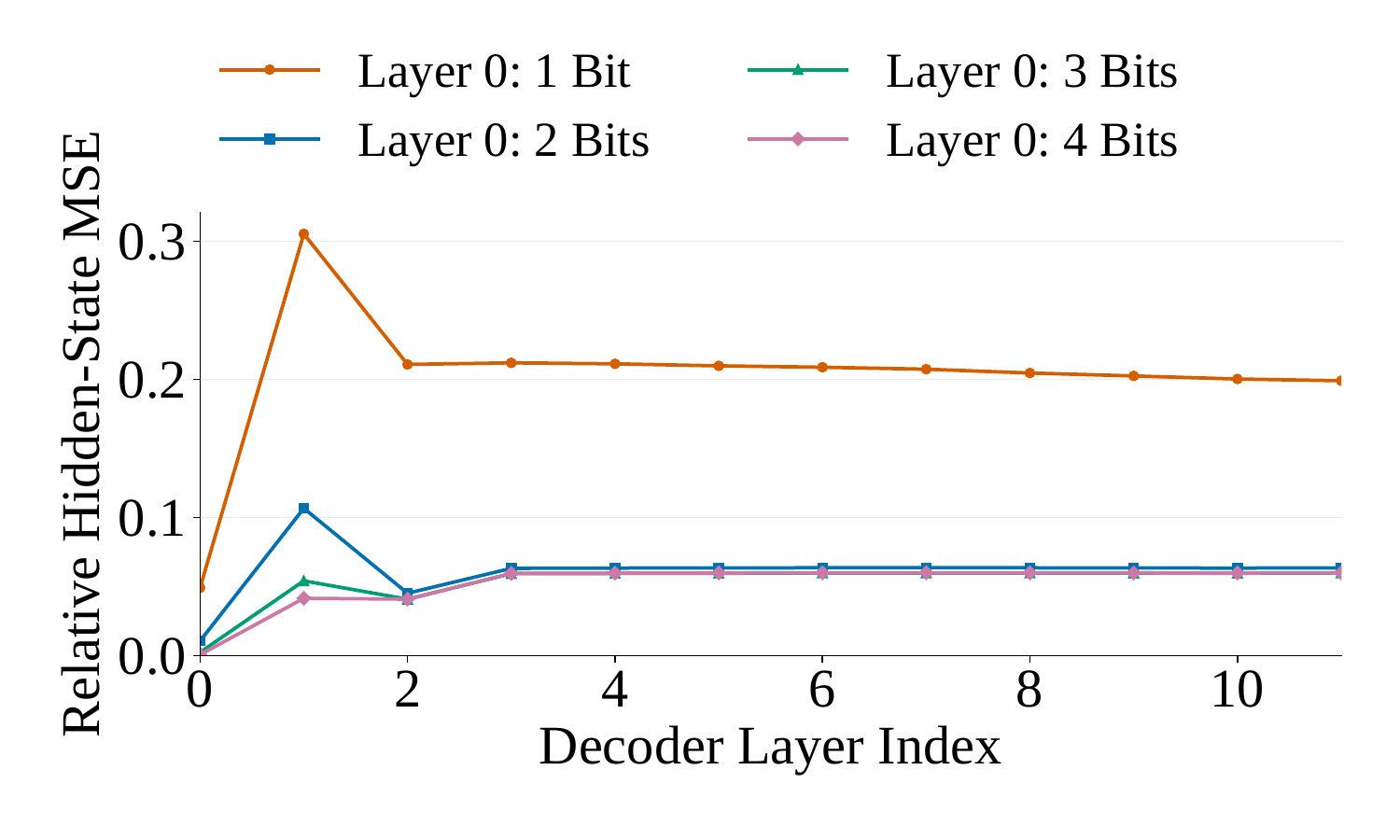}
    \vspace{-6mm}
    \caption{Relative hidden-state MSE at each decoder layer, measured against the corresponding full-precision model outputs. All experts in layer 0 are quantized to 1–4 bits, while those in other layers use 2 bits.}
\label{fig:relative_hidden_states_mse}
\vspace{-4mm}
\end{figure}

Existing MoE quantization studies~\citep{qin2026veq,hu2025moequant} mainly improve model performance through mixed-precision allocation~\citep{huang2025mixture,duanmu2025mxmoe,xie-etal-2025-automated,deng2026gemq} and post-quantization fine-tuning~\citep{deng2026gemq, chen2025eac}. The former assigns different bit-widths to experts based on expert sensitivity~\citep{chitty2025mopeq,xie-etal-2025-automated}, routing information~\citep{huang2025mixture,duanmu2025mxmoe}, or quantization error~\citep{deng2026gemq,duanmu2025mxmoe,qi-etal-2026-profiling}, while the latter adapts to the post-quantization distribution by updating the remaining full-precision parameters. However, these approaches still face two key challenges. First, existing mixed-precision allocation strategies typically rely on manually designed importance metrics and consecutive bit-width search spaces. Although these metrics can capture local quantization errors, our statistical experiments demonstrate that they may still underestimate how extremely low-bit quantization affects model performance (see Fig.~\ref{fig:relative_hidden_states_mse}). Second, although existing router fine-tuning methods can reduce perplexity on calibration data, such improvements do not consistently translate into performance gains on real-world downstream generation tasks.

To overcome these challenges, we propose \textbf{ProQuant}, an efficient PTQ framework for large-scale MoE models. First, we introduce Prune-or-Prioritize Mixed-Precision Allocation (PoP), which jointly determines expert pruning and bit-width assignment within a unified optimization. PoP allows low-contribution experts to be removed while prioritizing sufficient precision for more valuable experts, thereby avoiding severe representation degradation caused by extremely low bit-widths. Second, we propose Routing-Consistent RMSNorm Finetune (RoCo), which compensates for expert quantization errors through mathematically equivalent transformations while preserving the original routing behavior. As illustrated in Fig.~\ref{fig:performance}, ProQuant consistently outperforms all competing methods across all six benchmarks, maintaining strong capabilities even on tasks where other methods suffer substantial degradation.

\begin{figure*}[t]
    \centering
    \includegraphics[width=\linewidth]{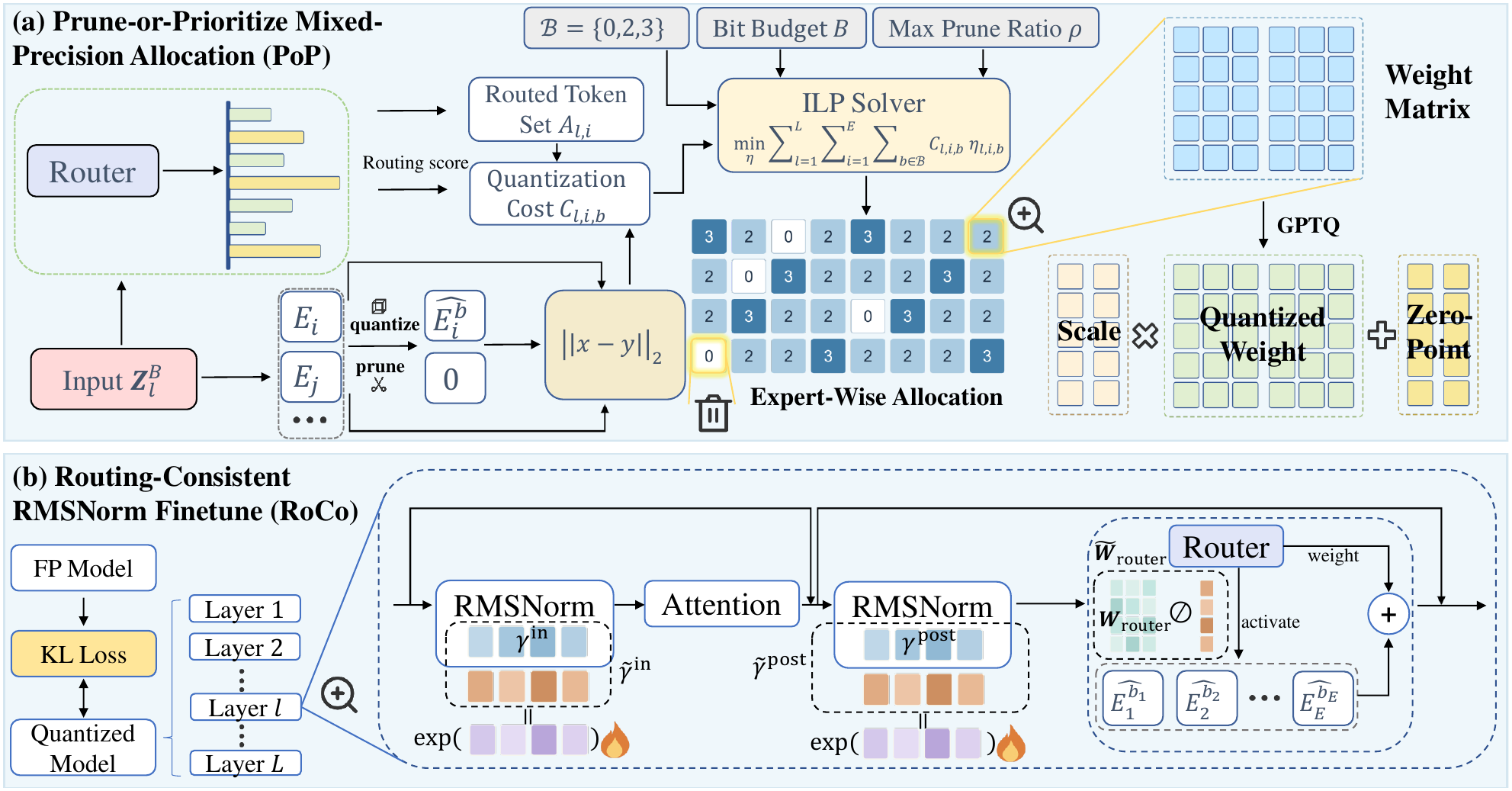}
    \caption{Overview of ProQuant, illustrating the workflows of \textbf{(a)} Prune-or-Prioritize Mixed-Precision Allocation (PoP) and \textbf{(b)} Routing-Consistent RMSNorm Finetune (RoCo).}
    \label{fig:overview}
    \vspace{-4mm}
\end{figure*}

Our contributions are summarized as follows:
\vspace{-1mm}
\begin{itemize}
\setlength{\itemsep}{1pt}
\item We propose \textbf{ProQuant}, a weight-only PTQ framework for large-scale MoE models that integrates mixed-precision allocation with lightweight post-quantization compensation.

\item We design \textbf{PoP}, which adopts a prune-or-prioritize joint decision to avoid severe degradation from extremely low-precision experts and better utilize limited bit budgets.

\item We develop \textbf{RoCo}, which decouples representation adaptation from routing behavior and compensates for quantization errors while preserving routing consistency. This avoids routing drift caused by directly fine-tuning full-precision parameters.

\item Experiments on Qwen3-30B-A3B-Instruct-2507 and Qwen3.5-35B-A3B validate ProQuant on diverse generative tasks at average expert precisions of 1.5, 2.0, and 2.5 bits.
\end{itemize}


\vspace{-2mm}
\section{Related Works}
\vspace{-1mm}

\subsection{Post-Training Quantization}

Post-training quantization (PTQ) directly converts pretrained models into low-precision representations without costly retraining. A representative direction is second-order error compensation, where GPTQ~\citep{frantar2022gptq} leverages Hessian information to compensate for quantization errors in the remaining weights during sequential column-wise quantization. Building on GPTQ, GPTAQ~\citep{li2025gptaq} and BoA~\citep{kim2024boa} extend second-order error compensation by accounting for inter-layer error accumulation and cross-projection dependencies, respectively. Activation-aware scaling is employed in AWQ~\citep{lin2024awq} to protect salient weight channels, whereas other studies~\citep{dettmers2023spqr,lee2024owq,kim2023squeezellm,shang2023pb,huang2024billm,huang2024slim} exploit differences in weight saliency or quantization sensitivity to adopt differentiated quantization strategies. Many PTQ methods use mathematically equivalent transformations to ease quantization. SmoothQuant~\citep{xiao2023smoothquant} adopts channel-wise scaling, while rotation-based methods~\citep{ashkboos2024quarot, liu2025spinquant, lin2024duquant} redistribute outliers. Other approaches~\citep{ma2024affinequant,shao2024omniquant,wei2023outlier} explore affine transformations and equivalent scaling or shifting operations. Recent studies have extended PTQ to ultra-low-bit regimes, including ternary, binary, and even sub-1-bit weight representations~\citep{yan2026ptllm, wang2026cat, chong2026nanoquant, li2025arb}. However, when applied to MoE architectures, these general-purpose PTQ methods often underutilize the bit budget, degrading performance.

\vspace{-2mm}
\subsection{Quantization of MoE Models}

MoE models pose additional quantization challenges due to sparse expert activation. To capture token–expert relevance, MoEQuant~\citep{hu2025moequant} and VEQ~\citep{qin2026veq} incorporate token–expert affinity into calibration and quantization objectives, with VEQ further introducing modality information for VLMs to account for textual and visual differences.
Router-aware methods~\citep{fu2025eaquant, chen2025eac, fang2026beyond} explicitly align the quantized routing behavior with its full-precision counterpart to preserve expert selection.
Expert heterogeneity also motivates mixed-precision quantization. Sensitivity-aware methods~\citep{li2024quantmoe, huang2025mixture, duanmu2025mxmoe, xie-etal-2025-automated} allocate precision according to structural or expert sensitivity. More recent studies develop principled sensitivity metrics~\citep{chowdhury2026efficient} or perform global expert-level bit allocation with router fine-tuning~\citep{deng2026gemq}. 
Beyond mixed-precision allocation, other works directly improve low-bit representations or compensate for quantization errors. MiLo~\citep{huang2025milo} reconstructs quantization residuals using adaptive low-rank compensators. CodeQuant~\citep{yin2026codequant} combines rotation and clustering to reduce quantization errors, whereas KBVQ-MoE~\citep{xu2026kbvq} removes cross-expert redundancy and applies bias correction.
Our method prunes less important experts to prioritize precision for others and preserves routing via router-decoupled RMSNorm fine-tuning.

\vspace{-2mm}
\section{Method}
\vspace{-1mm}
\subsection{Preliminaries}

\vspace{-1mm}
\textbf{Mixture-of-Experts.} MoE-based large language models~\citep{liu2024deepseek,xu2026deepseek,yang2025qwen3} replace each FFN layer of a dense model with an MoE layer consisting of a router and multiple parallel experts, such that only a subset of experts is activated during forward propagation. Given the input hidden state $\mathbf{z}$ of the $t$-th token to the $l$-th MoE layer, the router computes routing scores over all experts and selects the top-$K$ experts for computation. The output of the MoE layer can be formulated as:
\begin{equation}
    \mathbf{y} = \sum_{i \in \mathrm{Top}\text{-}K} g_i(\mathbf{z}) E_i(\mathbf{z}),
\end{equation}
where $E_i$ denotes the $i$-th expert, and $\mathrm{Top}\text{-}K$ is the index set of top-$K$ experts with the highest routing scores. The routing weight of expert $i$ is obtained by renormalizing its routing probability within $\mathrm{Top}\text{-}K$:
\begin{equation}
    g_i(\mathbf{z}) = \frac{\pi_i(\mathbf{z})}{\sum_{j \in \mathrm{Top}\text{-}K}\pi_j(\mathbf{z})},
\end{equation}
where $\pi_i(\mathbf{z})$ denotes the routing probability assigned to expert $i$.

\textbf{Weight-Only Quantization.}
Asymmetric quantization maps floating-point weights to low-bit integer representations, with a scaling factor and a zero point stored for each weight group. Given a weight matrix $\mathbf{W}\in\mathbb{R}^{C_{\mathrm{out}}\times C_{\mathrm{in}}}$, each row is partitioned into multiple groups. For each group $\mathbf{w}\in\mathbb{R}^{1\times c}$, its low-bit integer representation is obtained via the round-to-nearest rule as follows:
\begin{equation}
    \mathbf{q} = \operatorname{clamp}(\lfloor \frac{\mathbf{w}}{s} \rceil + z, q_{\min}, q_{\max}),
\end{equation}
where $s$ and $z$ denote the scaling factor and zero point, respectively, $[q_{\min},q_{\max}]$ denotes the representable range of the quantized integer, and $\lfloor \cdot \rceil$ denotes the rounding operation. The corresponding quantized approximation of the original weights can be expressed as:
\begin{equation}
    \hat{\mathbf{w}} = s(\mathbf{q} - z).
\end{equation}

GPTQ~\citep{frantar2022gptq} has demonstrated relatively robust performance under low-bit quantization. Therefore, once the target bit-width of a weight matrix is determined, we adopt it for quantization. Specifically, GPTQ minimizes the following layer-wise reconstruction loss:
\begin{equation}
    \min\|\mathbf{WX} - \mathbf{\hat{W}X}\|_F^2,
\end{equation}
where $\mathbf{X}$ denotes the calibration activations and $\hat{\mathbf{W}}$ the quantized weights. It uses the Hessian approximation $\mathbf{H}=2\mathbf{X}\mathbf{X}^{\top}$ to guide quantization error compensation, helping preserve reconstruction quality at low precision.

\vspace{-4mm}
\subsection{Prune-or-Prioritize Mixed-Precision Allocation}
\begin{figure}[t]
    \centering
    \includegraphics[width=\linewidth]{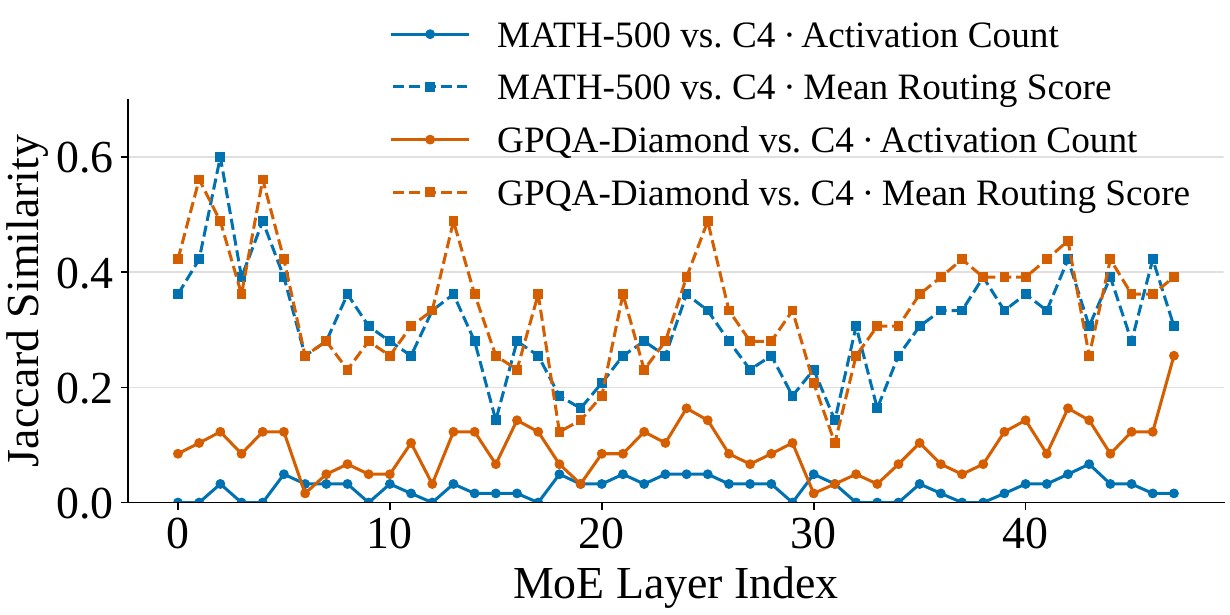}
    \vspace{-6mm}
    \caption{Top-32 expert Jaccard similarity across MoE layers in Qwen3-30B-A3B-Instruct-2507, comparing C4 with MATH-500 and GPQA-Diamond using activation-count and mean-routing-score rankings.}
    \label{fig:diff_ds_jaccard}
    \vspace{-6mm}
\end{figure}

\vspace{-2mm}
To quantify the quantization error of an expert, we introduce the definition of \textbf{expert quantization cost}. As MoE outputs are weighted by routing scores, we incorporate their renormalized values into its definition:
\begin{equation}
\resizebox{0.89\columnwidth}{!}{$\displaystyle
C_{l,i,b} =
\frac{
\sum_{t\in A_{l,i}} g_{l,i}(\mathbf{z}_{t,l}^{B})
\|E_{l,i}(\mathbf{z}_{t,l}^{B})
-\hat{E}_{l,i}^{b}(\mathbf{z}_{t,l}^{B})\|_2
}{|A_{l,i}|},
$}
\end{equation}
where $C_{l,i,b}$ represents the quantization cost of expert $(l,i)$ at $b$ bits, with $l$ and $i$ indexing the MoE layer and the expert within that layer. $E_{l,i}$ and $\hat{E}_{l,i}^{b}$ correspond to the full-precision expert and its $b$-bit quantized counterpart. $\mathbf{z}_{t,l}^{B}$ is obtained by propagating token $t$ through the preceding layers quantized to an average bit-width of $B$. $A_{l,i}$ contains the indices of tokens routed to expert $i$ in layer $l$. As shown in Fig.~\ref{fig:diff_ds_jaccard}, we compare the similarity of the top-32 experts ranked by activation count and mean routing score between C4 and two downstream benchmarks. Mean routing score yields more consistent expert selections across datasets, suggesting that it better reflects stable expert importance. We therefore use routing-score-weighted quantization error and normalize it by $|A_{l,i}|$ to remove the effect of activation frequency.

As shown in Fig.~\ref{fig:relative_hidden_states_mse}, we quantize all experts in layer 0 to 1, 2, 3, or 4 bits, while keeping subsequent layers at 2 bits, to investigate cross-layer error propagation. Compared with other bit-widths, the errors introduced by 1-bit quantization cannot be effectively compensated for by subsequent higher-precision layers. This suggests that extremely low-bit quantization can cause persistent degradation in hidden representations across layers. However, some existing methods~\citep{huang2025mixture,duanmu2025mxmoe} estimate quantization sensitivity using local output errors, potentially underestimating such cross-layer effects and the actual impact of extremely low precision. Based on this observation, we exclude 1-bit quantization from the candidate bit-widths under average expert bit-width budgets of 1.5, 2.0, and 2.5 bits. Instead, we introduce a 0-bit pruning option, whose expert quantization cost is defined as follows:
\begin{equation}
    C_{l,i,0} = \frac{\sum_{t\in A_{l,i}}g_{l,i}(\mathbf{z}_{t,l}^B)\|E_{l,i}(\mathbf{z}_{t,l}^B)\|_2}{|A_{l,i}|},
\end{equation}
where 0-bit means pruning the expert. Replacing 1-bit with 0-bit brings two benefits: (1) it frees additional bit budget for more important experts, allowing them to retain higher precision; (2) it removes severely compressed experts with limited representational capacity and allows higher-precision experts to take their place, which better exploits the sparse activation property of MoE models.

Based on the above definitions, we formulate the bit-width allocation as the following integer linear programming (ILP) problem:
\begin{equation}
\label{eq:ILP}
\begin{aligned}
\min_{\eta} \quad & \sum_{l=1}^{L} \sum_{i=1}^{E} \sum_{b\in\mathcal B} C_{l,i,b} \, \eta_{l,i,b} \\
\text{s.t.} \quad
& \eta_{l,i,b} \in \{0, 1\}, \; \forall l,i,b \\
& \sum_{b\in\mathcal B} \eta_{l,i,b} = 1, \; \forall l,i \\
& \sum_{i=1}^{E} \eta_{l,i,0} \le \min\{E - K, \rho E\}, \; \forall l \\
& \sum_{l=1}^L \sum_{i=1}^E \sum_{b\in\mathcal B} b \, \eta_{l,i,b} \le BLE,
\end{aligned}
\end{equation}
where $\eta_{l,i,b}$ indicates whether expert $(l,i)$ is quantized to $b$ bits, and $\mathcal{B}$ denotes the set of candidate bit-widths, which is set to $\{0,2,3\}$ in our settings. To avoid excessive pruning, we further bound the number of $0$-bit experts by the maximum ratio $\rho$. Solving this ILP jointly determines expert pruning and the bit-widths of retained experts.


\vspace{-2mm}
\subsection{Routing-Consistent RMSNorm Finetune}
Prior studies either calibrate the router through routing alignment~\citep{fu2025eaquant,chen2025eac}, or adapt it after quantization using a global output-level objective~\citep{deng2026gemq}. However, we find that neither strategy yields consistent improvements on downstream tasks. Routing alignment constrains the quantized model to match the routing decisions of its full-precision counterpart. However, such local alignment lacks global information about the quantized model. Therefore, it cannot effectively adjust routing to compensate for quantization errors accumulated across layers. In contrast, directly optimizing the router with a global output-level objective can adapt routing according to the quantized model. However, fitting the router to the calibration set may bias it toward experts that are more frequently activated by the calibration data. Inspired by~\citep{xiao2023smoothquant,lin2024awq}, we propose an RMSNorm fine-tuning method based on mathematically equivalent transformations. It adapts hidden representations for quantization-error compensation while preserving the original routing behavior.

Given an input hidden state $\mathbf{x}\in\mathbb{R}^{1\times d}$ to the RMSNorm layer following the attention module, its output is given by
\begin{equation}
    \mathbf{z} = \boldsymbol{\gamma} \; \odot \; \operatorname{Norm}(\mathbf{x}),
\end{equation}
where $\boldsymbol{\gamma}\in\mathbb{R}^{1\times d}$ denotes the learnable scaling parameter, and $\operatorname{Norm}(\mathbf{x}) = \frac{\mathbf{x}}{\sqrt{\frac{1}{d}\mathbf{x}\mathbf{x}^{\top} + \epsilon}}$.

\begin{figure}[t]
    \centering
    \includegraphics[width=\linewidth]{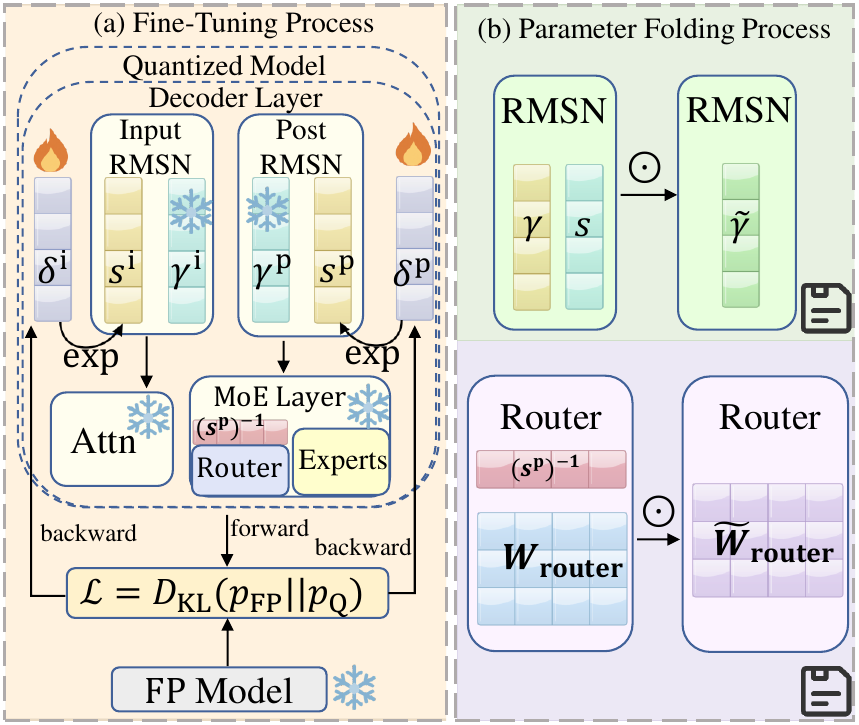}
    \vspace{-3mm}
    \caption{Architecture of RoCo. (a) Fine-tuning process and (b) parameter folding process. Superscripts $i$ and $p$ denote “in” and “post”, respectively.}
    \label{fig:roco}
    \vspace{-6mm}
\end{figure}

\begin{figure*}[t]
    \centering
    \includegraphics[width=\linewidth]{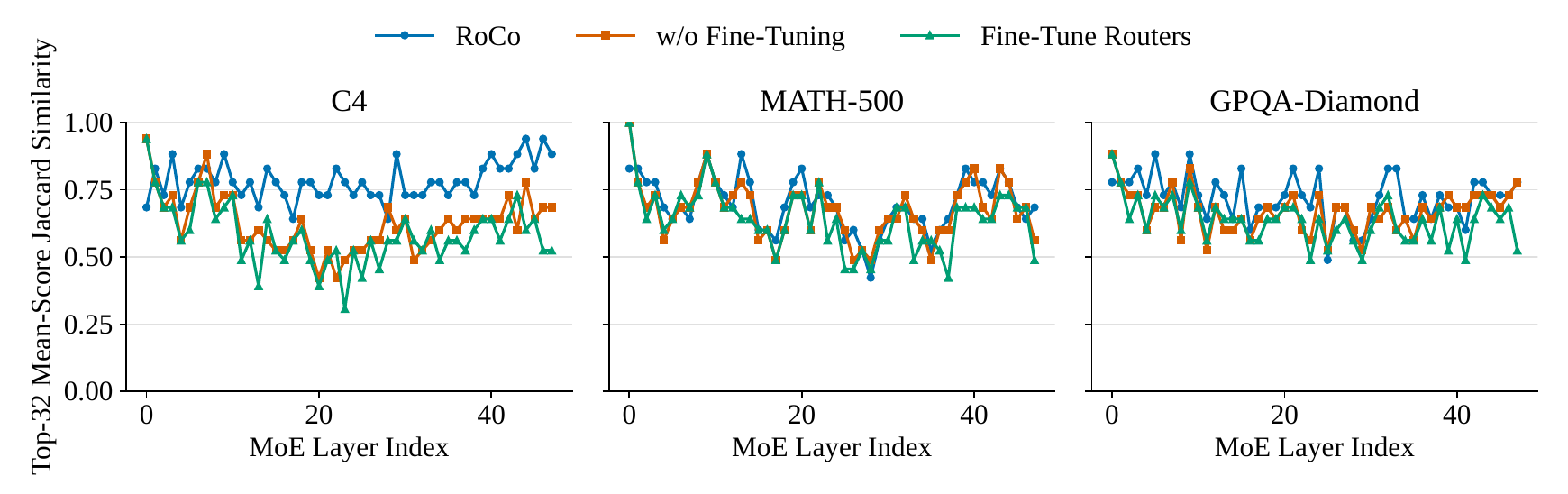}
    \vspace{-6mm}
    \caption{Layer-wise Jaccard similarity between the full-precision and quantized models’ top-32 experts ranked by mean routing score on three datasets.}
    \label{fig:finetune_jaccard}
    \vspace{-5mm}
\end{figure*}

We introduce $\boldsymbol{\delta}\in\mathbb{R}^{1\times d}$ as the optimization variable and parameterize a positive scaling vector as $\mathbf{s}=\exp(\boldsymbol{\delta})$. The scaling vector is applied to the learnable RMSNorm parameter $\boldsymbol{\gamma}$, while the corresponding input channels of the router weight matrix are inversely scaled to preserve the routing outputs:
\begin{equation}
    \widetilde{\boldsymbol{\gamma}} = \boldsymbol{\gamma}\odot\mathbf{s},\;
    \widetilde{\mathbf{W}}_{\mathrm{router}} = \mathbf{W}_{\mathrm{router}} \odot (\mathbf{1}^{\top}\mathbf{s}^{-1}),
\end{equation}
where $\odot$ denotes element-wise multiplication, $\mathbf{1}\in\mathbb{R}^{1\times C_{\mathrm{out}}}$ is an all-ones vector, and $\mathbf{s}^{-1}$ denotes the element-wise reciprocal of $\mathbf{s}$. After applying the scaling, the RMSNorm output becomes $\widetilde{\mathbf{z}} = \widetilde{\boldsymbol{\gamma}} \odot \operatorname{Norm}(\mathbf{x}) = \mathbf{z}\odot\mathbf{s}$.
The inverse scaling on $\mathbf{W}_{\mathrm{router}}$ exactly cancels this change when computing the router logits:
\begin{equation}
    \widetilde{\mathbf{z}}
    \widetilde{\mathbf{W}}_{\mathrm{router}}^{\top}
    =
    \mathbf{z}
    \mathbf{W}_{\mathrm{router}}^{\top}.
\end{equation}
Therefore, the equivalent transformation allows RMSNorm fine-tuning to compensate for expert quantization errors while preserving the original routing behavior, thereby avoiding routing flips induced by RMSNorm updates. We optimize the RMSNorm with the following KL divergence:
\begin{equation}
\mathcal{L} = D_{\mathrm{KL}}(\mathbf{p}_{\mathrm{FP}}\|\mathbf{p}_{\mathrm{Q}})
\end{equation}
where $\mathbf{p}_{\mathrm{FP}}$ and $\mathbf{p}_{\mathrm{Q}}$ denote the output probability distributions of the full-precision and quantized models, respectively. Since only a small number of RMSNorm weights are updated, the fine-tuning is highly lightweight. Moreover, it consistently outperforms direct fine-tuning without inverse scaling of the router weights, as shown in ablation study. After this step, the learned scaling vector is absorbed into the RMSNorm and router weights, introducing no additional overhead during inference. The process is illustrated in Fig.~\ref{fig:roco}.

Fig.~\ref{fig:finetune_jaccard} reports the Jaccard similarity of top-32 expert sets ranked by mean routing score between the full-precision model and the same quantized model under three strategies. RoCo achieves higher Jaccard similarity across most MoE layers, indicating that it better preserves the expert-selection preference of the full-precision model for the target task. By maintaining the original routing behavior, RoCo allows the quantized model to continue prioritizing experts favored by the full-precision model, while adapting hidden representations to compensate for quantization errors. RoCo stabilizes routing and strengthens experts.


\begin{table*}[t]
\centering
\small
\setlength{\tabcolsep}{5pt}

\vspace{-2mm}

\begin{tabular}{>{\centering\arraybackslash}p{1.6cm}|c|c|cccccc|c}
\hline
\textbf{\centering Model} & \textbf{Bits} & \textbf{Method} & \textbf{GSM8K} & \textbf{MATH-500} & \textbf{GPQA-D} & \textbf{IFEval} & \textbf{MMLU-R} & \textbf{BFCL-v3} & \textbf{AVG} \\
\hline

\multirow{14}{*}{\centering\parbox{1.6cm}{\centering Qwen3-30B-A3B-\\Instruct-2507}}
& 16.0 & BF16 & 96.36 & 95.40 & 55.05 & 87.66 & 88.18 & 63.02 & 80.94 \\
\cline{2-10}
& \multirow{4}{*}{2.5} & MxMoE & 93.86 & 51.80 & 36.36 & 81.17 & 81.26 & 30.01 & 62.41 \\
& & PMQ & 95.45 & 39.80 & 30.81 & 82.33 & 79.56 & 29.82 & 59.63 \\
& & GEMQ & 93.93 & 45.60 & 31.31 & 82.66 & 81.56 & 30.22 & 60.88 \\
& & ProQuant & \textbf{95.53} & \textbf{93.20} & \textbf{53.03} & \textbf{87.19} & \textbf{85.44} & \textbf{56.31} & \textbf{78.45} \\
\cline{2-10}
& \multirow{5}{*}{2.0} & GPTQ & 91.28 & 76.00 & 36.36 & 81.17 & \textbf{81.75} & 45.29 & 68.64 \\
& & MxMoE & 81.43 & 15.60 & 21.72 & 75.80 & 74.65 & 30.11 & 49.89 \\
& & PMQ & 84.61 & 12.80 & 24.24 & 75.03 & 71.44 & 29.75 & 49.65 \\
& & GEMQ & 84.31 & 18.20 & 24.24 & 73.95 & 71.75 & 30.08 & 50.42 \\
& & ProQuant & \textbf{94.47} & \textbf{90.20} & \textbf{45.96} & \textbf{84.47} & 79.82 & \textbf{52.05} & \textbf{74.49} \\
\cline{2-10}
& \multirow{4}{*}{1.5} & MxMoE & 5.38 & 3.60 & 13.64 & 39.93 & 55.96 & 29.61 & 24.69 \\
& & PMQ & 1.44 & 1.40 & 19.19 & 36.51 & 44.26 & 29.73 & 22.09 \\
& & GEMQ & 29.72 & 2.80 & 22.22 & 46.90 & 53.14 & 29.73 & 30.75 \\
& & ProQuant & \textbf{93.93} & \textbf{86.00} & \textbf{32.83} & \textbf{77.07} & \textbf{67.26} & \textbf{46.69} & \textbf{67.30} \\

\hline
\multirow{14}{*}{\centering\parbox{1.6cm}{\centering Qwen3.5-35B-A3B}}
& 16.0 & BF16 & 97.04 & 93.60 & 81.82 & 90.18 & 92.39 & 68.87 & 87.32 \\
\cline{2-10}
& \multirow{4}{*}{2.5} & MxMoE & 96.36 & 81.80 & 56.06 & 87.95 & 90.04 & \textbf{63.13} & 79.22 \\
& & PMQ & 96.29 & 71.20 & 39.39 & 86.82 & 86.53 & 56.80 & 72.84 \\
& & GEMQ & 95.68 & 67.40 & 45.45 & 85.51 & 87.77 & 45.23 & 71.17 \\
& & ProQuant & \textbf{96.89} & \textbf{93.60} & \textbf{70.20} & \textbf{88.03} & \textbf{91.33} & 60.75 & \textbf{83.47} \\
\cline{2-10}
& \multirow{5}{*}{2.0} & GPTQ & 93.10 & \textbf{83.00} & 53.54 & 82.83 & 88.44 & \textbf{59.79} & 76.78 \\
& & MxMoE & 85.60 & 47.20 & 27.78 & 79.22 & 82.30 & 40.79 & 60.48 \\
& & PMQ & 94.31 & 42.60 & 26.26 & 77.01 & 77.09 & 45.12 & 60.40 \\
& & GEMQ & 72.40 & 29.80 & 24.24 & 75.37 & 78.91 & 48.13 & 54.81 \\
& & ProQuant & \textbf{95.53} & 82.60 & \textbf{57.07} & \textbf{87.76} & \textbf{89.05} & 57.66 & \textbf{78.28} \\
\cline{2-10}
& \multirow{4}{*}{1.5} & MxMoE & 69.29 & 17.60 & 22.73 & 47.05 & 69.28 & 22.57 & 41.42 \\
& & PMQ & 22.29 & 6.60 & 14.65 & 48.46 & 58.54 & 17.96 & 28.08 \\
& & GEMQ & 36.92 & 11.80 & 27.27 & 55.46 & 64.88 & 12.03 & 34.73 \\
& & ProQuant & \textbf{94.84} & \textbf{40.20} & \textbf{34.85} & \textbf{81.16} & \textbf{81.72} & \textbf{32.31} & \textbf{60.85} \\

\hline
\end{tabular}
\caption{Performance of quantized Qwen3-30B-A3B-Instruct-2507 and Qwen3.5-35B-A3B on six generation-based benchmarks. We compare ProQuant with representative mixed-precision methods for MoE LLMs at varying average expert bit-widths. GPQA-D and MMLU-R denote GPQA-Diamond and MMLU-Redux, respectively.}
\label{tab:main_results}
\end{table*}

\vspace{-2mm}
\section{Experiments}
\vspace{-1mm}
\subsection{Settings}
\vspace{-1mm}
\textbf{Implementation Details.} All experiments are conducted using PyTorch~\citep{paszke2019pytorch} and HuggingFace Transformers~\citep{wolf-etal-2020-transformers} on two NVIDIA A800 GPUs. We evaluate our method on Qwen3-30B-A3B-Instruct-2507~\citep{yang2025qwen3} and Qwen3.5-35B-A3B~\citep{qwen3.5}. We use 128 sequences of 2,048 tokens sampled from C4~\citep{raffel2020exploring} with seed 0 as the calibration set. For both softmax and linear attention layers, the Q, K, V, and O projection weights are uniformly quantized to 4 bits, while all other linear-attention weights and router weights remain in full precision. We consider average expert bit-widths of 2.5, 2.0, and 1.5 bits with a candidate set of ${0,2,3}$ bits, and quantize the shared experts of Qwen3.5 to 4 bits. See Appendix~\ref{app:implementations} for details.

\textbf{Models and Evaluation.} We observe that the conventional zero-shot evaluation with lm-eval~\citep{biderman2024lessons}, which is widely adopted in prior work, may not fully reflect model capability on practical generative tasks. Therefore, we primarily use EvalScope~\citep{evalscope_2024} to evaluate generation-based performance across diverse capabilities, such as mathematical reasoning, scientific reasoning, and instruction following, on six generative benchmarks: GPQA-Diamond~\citep{rein2024gpqa}, MATH-500~\citep{hendrycks2021measuring,lightman2024lets}, IFEval~\citep{zhou2023instruction}, GSM8K~\citep{cobbe2021training}, MMLU-Redux~\citep{gema2025we} and BFCL-v3~\citep{pmlr-v267-patil25a}. We also report perplexity and zero-shot results as auxiliary metrics.

\textbf{Baselines.} We compare ProQuant with representative mixed-precision quantization methods for MoE LLMs, including PMQ from MC-MoE~\citep{huang2025mixture}, MxMoE~\citep{duanmu2025mxmoe}, and GEMQ~\citep{deng2026gemq}. For fair comparison, all methods use GPTQ~\citep{frantar2022gptq} with their respective bit-width allocation or post-quantization compensation strategies.

\vspace{-1mm}
\subsection{Main Results}
\vspace{1mm}

\begin{table*}[t]
\centering
\small
\setlength{\tabcolsep}{4pt}

\begin{tabular}{>{\centering\arraybackslash}p{2.5cm}|c|c|cc|ccccc|c}
\hline
\textbf{Model} & 
\textbf{Bits} & 
\textbf{Method} & 
\textbf{Wiki. PPL}$\downarrow$ & 
\textbf{C4 PPL}$\downarrow$ & 
\textbf{PIQA}$\uparrow$ & 
\textbf{ARC-E}$\uparrow$ & 
\textbf{ARC-C}$\uparrow$ & 
\textbf{HS}$\uparrow$ & 
\textbf{WING}$\uparrow$ & 
\textbf{AVG}$\uparrow$ \\
\hline
\multirow{6}{*}{\parbox{2.5cm}{\centering Qwen3-30B-A3B-Instruct-2507}} 
& 16.0 & BF16 & 7.26 & 12.49 & 80.09 & 83.88 & 62.37 & 79.72 & 73.40 & 75.89 \\
\cline{2-11}
& \multirow{5}{*}{2.0} & GPTQ & 9.83 & 15.45 & 78.29 & \textbf{80.18} & 56.83 & 73.70 & 71.74 & 72.15 \\
& & MxMoE & 9.23 & 13.97 & 79.54 & 79.21 & 55.20 & \textbf{77.84} & 72.30 & \textbf{72.82} \\
& & PMQ & 10.61 & 14.84 & \textbf{79.71} & 78.87 & 54.69 & 77.73 & \textbf{72.93} & 72.79 \\
& & GEMQ & \textbf{8.22} & \textbf{13.49} & 79.65 & 77.99 & 56.31 & 76.97 & 72.61 & 72.79 \\
& & ProQuant & 8.94 & 14.57 & 79.22 & 79.08 & \textbf{57.17} & 75.19 & 70.64 & 72.26 \\
\hline
\end{tabular}
\caption{Perplexity and 0-shot log-likelihood accuracy on Qwen3-30B-A3B-Instruct-2507. Wiki., ARC-E, ARC-C, HS, and WING denote WikiText-2, ARC-Easy, ARC-Challenge, HellaSwag, and WinoGrande, respectively.}
\label{tab:ppl_zero}
\vspace{-4mm}
\end{table*}

\vspace{-2mm}
\begin{table*}[t]
\centering
\small
\setlength{\tabcolsep}{6pt}
\begin{tabular}{c|cccccc|c}
\hline
\textbf{Candidate Bits} & \textbf{GSM8K} & \textbf{MATH-500} & \textbf{GPQA-D} & \textbf{IFEval} & \textbf{MMLU-R} & \textbf{BFCL-v3} & \textbf{AVG} \\
\hline
$\{0,2,3\}$   & \textbf{94.47} & \textbf{90.20} & \textbf{45.96} & \textbf{84.47} & \textbf{79.82} & \textbf{52.05} & \textbf{74.49} \\
$\{1,2,3\}$   & 93.33 & 85.80 & 36.87 & 80.96 & 71.40 & 42.40 & 68.46 \\
$\{0,1,2,3\}$ & 93.93 & 85.40 & 36.36 & 77.88 & 69.44 & 38.87 & 66.98 \\
\hline
\end{tabular}
\vspace{-2mm}
\caption{Ablation of candidate bit-width sets on Qwen3-30B-A3B-Instruct-2507 at a 2.0-bit average expert precision.}
\label{tab:ablation_pruning}
\end{table*}

\begin{table*}[t]
\centering
\small
\setlength{\tabcolsep}{6pt}
\begin{tabular}{c|cccccc|c}
\hline
\textbf{Method} & \textbf{GSM8K} & \textbf{MATH-500} & \textbf{GPQA-D} & \textbf{IFEval} & \textbf{MMLU-R} & \textbf{BFCL-v3} & \textbf{AVG} \\
\hline
ProQuant          & 93.93 & 86.00 & \textbf{32.83} & \textbf{77.07} & \textbf{67.26} & \textbf{46.69} & \textbf{67.30} \\
w/o RoCo      & \textbf{94.16} & 85.80 & 26.77 & 75.35 & 63.75 & 42.79 & 64.77 \\
Direct RMSNorm FT  & 93.33 & \textbf{86.60} & 30.81 & 74.96 & 65.88 & 43.89 & 65.91 \\
\hline
\end{tabular}
\vspace{-2mm}
\caption{Fine-tuning strategy ablation on Qwen3-30B-A3B-Instruct-2507 at a 1.5-bit average expert precision.}
\label{tab:ablation_finetune}
\end{table*}

\begin{table*}[!h]
\centering
\small
\setlength{\tabcolsep}{5pt}

\begin{tabular}{>{\centering\arraybackslash}p{1.6cm}|c|c|cccccc|c}
\hline
\textbf{Model} & \textbf{Bits} & $\boldsymbol{\rho}$ & \textbf{GSM8K} & \textbf{MATH-500} & \textbf{GPQA-D} & \textbf{IFEval} & \textbf{MMLU-R} & \textbf{BFCL-v3} & \textbf{AVG} \\
\hline

\multirow{5}{*}{\centering\parbox{1.6cm}{\centering Qwen3-30B-A3B-\\Instruct-2507}}
& \multirow{5}{*}{2.0} 
& 0.10 & 95.00 & 88.00 & \textbf{48.99} & 82.10 & \textbf{82.56} & 49.71 & 74.39 \\
& & 0.15 & 94.69 & 88.20 & 42.93 & 84.21 & 81.30 & 52.66 & 74.00 \\
& & 0.20 & 95.22 & 87.80 & 44.44 & \textbf{84.92} & 80.12 & 51.67 & 74.03 \\
& & 0.25 & \textbf{95.45} & \textbf{91.80} & 46.46 & 82.57 & 79.49 & \textbf{54.38} & \textbf{75.03} \\
& & 0.30 & 94.24 & 90.60 & 45.45 & 83.97 & 79.00 & 51.04 & 74.05 \\

\hline
\end{tabular}
\vspace{-2mm}
\caption{Sensitivity to $\rho$ on Qwen3-30B-A3B-Instruct-2507 with 2.0-bit experts, without RoCo finetuning.}
\label{tab:rho_ablation}
\vspace{-2mm}
\end{table*}

\textbf{Generation-based Evaluation.}
We first evaluate the performance of quantized models on realistic generation tasks. Table~\ref{tab:main_results} shows that ProQuant consistently achieves the best or competitive performance across different models and bit-width budgets, with more pronounced advantages under low-bit settings. As the expert average bit-width decreases from 2.5 bits to 1.5 bits, existing methods suffer from substantial performance degradation, while ProQuant maintains strong generation capability. For example, under the 1.5-bit setting, ProQuant achieves average scores of 67.30 and 60.85 on Qwen3-30B-A3B-Instruct-2507 and Qwen3.5-35B-A3B, respectively, significantly outperforming existing mixed-precision quantization methods. These results demonstrate that ProQuant can better preserve the capabilities of MoE LLMs under extremely constrained bit budgets.

\textbf{Perplexity and Log-Likelihood Evaluation.}
Furthermore, following previous works, we report perplexity on WikiText-2 and C4, and conduct zero-shot likelihood-based evaluation using lm-eval in Table~\ref{tab:ppl_zero}. The results show that ProQuant improves performance on realistic generation tasks while maintaining comparable results on conventional likelihood-based benchmarks. However, when considered together with Table~\ref{tab:main_results}, these results suggest that strong performance on PPL and zero-shot likelihood-based metrics does not necessarily translate into better generation ability. In particular, some mixed-precision quantization methods achieve favorable results on traditional likelihood-based evaluations, yet perform worse than a uniformly quantized 2.0-bit baseline on generation-based tasks. This indicates that token-level likelihood estimation, which underlies these conventional metrics, cannot fully reflect the practical capabilities of quantized models in open-ended generation scenarios and may lead to biased conclusions about the effectiveness of MoE quantization methods. Therefore, we argue that generation-based evaluation is essential for a comprehensive assessment of MoE quantization models.

\vspace{-2mm}
\subsection{Ablation Study}

\vspace{-1mm}

\textbf{Effect of PoP.} PoP adopts a fixed candidate bit-width set of $\{0,2,3\}$. To validate this design choice, we construct two ablation variants: $\{1,2,3\}$ removes the 0-bit option and forces all experts to be retained, while $\{0,1,2,3\}$ additionally introduces 1-bit quantization. The full PoP setting achieves the highest average score of 74.07, compared with 68.46 and 66.98 for the two variants, respectively. Notably, introducing the 0-bit option does not make $\{0,1,2,3\}$ outperform $\{1,2,3\}$, indicating that the gain does not come from pruning itself, but from avoiding severely quantized 1-bit experts that continue to participate in the forward pass. Thanks to the sparse routing mechanism of MoE models, a removed expert can be replaced by a slightly lower-ranked expert that retains higher precision. PoP prunes experts impaired by extremely low precision and reallocates their bit budget to preserve others’ effectiveness.

\textbf{Effect of RoCo.} We evaluate the effectiveness of RoCo by comparing the full ProQuant framework with two ablation variants: w/o RoCo, which completely removes RoCo, and Direct RMSNorm FT, which directly fine-tunes RMSNorm without preserving routing consistency. The average score drops from 67.30 to 64.77 without RoCo, indicating that lightweight RMSNorm fine-tuning can effectively mitigate quantization-induced performance degradation. Direct RMSNorm FT improves the average score to 65.91, but still underperforms the full ProQuant framework. This suggests that directly updating RMSNorm can provide partial recovery, but may also alter the router input and disturb the original routing behavior. RoCo decouples representation adaptation from routing, correcting quantization errors while preserving expert selection and improving recovery on generative tasks.

\textbf{Sensitivity to the Pruning Ratio.} Since PoP and RoCo are decoupled, we analyze the effect of the maximum pruning ratio $\rho$ without RoCo fine-tuning, so as to isolate the sensitivity of PoP to this hyperparameter. On Qwen3-30B-A3B-Instruct-2507, the average performance varies only slightly as $\rho$ ranges from 0.10 to 0.30, indicating that PoP is relatively robust to $\rho$ within this range on this model. When $\rho$ exceeds 0.30, the constraint is no longer active, and further increasing it does not affect the final allocation. The results also show that a moderate maximum pruning ratio is generally favorable for achieving higher average accuracy.


\vspace{-2mm}
\section{Conclusion}
\vspace{-2mm}

\label{sec:bibtex}

In this work, we presented ProQuant, which preserves MoE LLM performance on downstream generation tasks. More importantly, our study offers two insights into MoE quantization. First, we advocate a principle for extremely low-bit mixed-precision allocation: \textbf{“discard it rather than destroy it.”} When the bit budget cannot adequately preserve an expert's functionality, removing it may be preferable to retaining a severely degraded version. Second, we caution against over-optimizing for calibration data or specific metrics and emphasize generalization to downstream tasks. This also calls for reexamining the use of perplexity and log-likelihood metrics: they measure a model's ability to assign probabilities to reference sequences, rather than to generate text autoregressively based on its own outputs, and cannot replace real-world generation evaluations. We hope these findings encourage MoE quantization research to prioritize model reliability and cross-task generalization.

\section*{Limitations}
In this work, we focus on mixed-precision PTQ for large-scale MoE LLMs and evaluate ProQuant on models with up to 35 billion total parameters. Although ProQuant demonstrates strong generation performance at this scale, resource constraints prevented us from extending our experiments to models with hundreds of billions or trillions of parameters. Its effectiveness and scalability at these larger scales therefore remain to be validated.


\bibliography{custom}

\clearpage
\appendix

\label{sec:appendix}

\section{ProQuant Pipeline and Algorithms}
As shown in Fig.~\ref{fig:overview}, ProQuant consists of three stages. First, as illustrated in Algorithm~\ref{alg:allocation}, we traverse the MoE layers of the pretrained model and evaluate the quantization cost of each expert under different candidate bit-widths, thereby determining the final bit-width of each expert under the global bit budget. We then apply GPTQ according to the resulting allocation to obtain the integer representations of the quantized weights, together with the corresponding scales and zero-points. Finally, as shown in Algorithm~\ref{alg:rmsnorm_ft}, we freeze all other model parameters and fine-tune the RMSNorm scaling vectors to compensate for quantization errors. After fine-tuning, the learned scaling vectors are fused into the corresponding RMSNorm weights, while their element-wise inverses are fused into the associated router weights, yielding the final ProQuant model.

\begin{algorithm}[t]
\caption{Prune-or-Prioritize Mixed-Precision Allocation}
\label{alg:allocation}
\begin{algorithmic}[1]

\Require MoE model $\mathcal{M}$, calibration set $\mathcal{D}$, average bit-width $B$, candidate bit-width set $\mathcal{B}$, maximum pruning ratio $\rho$, top-$K$ value $K$
\Ensure Bit-width assignment $\boldsymbol{\eta}$

\State $\mathbf{Z}_{1}^{B} \gets \Call{GetMoEInput}{\mathcal{M}, \mathcal{D}}$

\For{$l = 1$ to $L$}

    \For{$i = 1$ to $E$}
        \State Obtain $A_{l,i}$ and $g_{l,i}$ using $\mathbf{Z}_{l}^{B}$

        \ForAll{$b \in \mathcal{B}$}
            \State $\hat{E}_{l,i}^{b} \gets \Call{Quantize}{E_{l,i}, b}$
            \State $C_{l,i,b} \gets \Call{CalcCost}{l,i,b,\mathbf{Z}_l^B}$
        \EndFor
    \EndFor

    \State $\hat{\mathcal{M}}_{l}^{B}
    \gets \Call{QuantizeLayer}{\mathcal{M}_{l}, B}$

    \State $\mathbf{Z}_{l+1}^{B}
    \gets \Call{LayerForward}{
        \hat{\mathcal{M}}_{l}^{B},
        \mathbf{Z}_{l}^{B}
    }$

\EndFor

\State $\boldsymbol{\eta}
\gets \Call{SolveILP}{
    C, B, \rho, K, \mathcal{B}, L, E
}$

\State \Return $\boldsymbol{\eta}$

\end{algorithmic}
\end{algorithm}

\begin{algorithm}[t]
\caption{Routing-Consistent RMSNorm Finetune}
\label{alg:rmsnorm_ft}
\begin{algorithmic}[1]

\Require Quantized model $\hat{\mathcal{M}}$, calibration set $\mathcal{D}$, learning rate $\alpha$, training steps $T$
\Ensure Fine-tuned quantized model $\hat{\mathcal{M}}$

\State Freeze all model weights
\State Initialize $\boldsymbol{\delta}_{l}^{\mathrm{in}} \gets \mathbf{0}$ and $\boldsymbol{\delta}_{l}^{\mathrm{post}} \gets \mathbf{0}, \; \forall l$

\For{$t = 1$ to $T$}

    \For{$l = 1$ to $L$}
        \State $\mathbf{s}_{l}^{\mathrm{in}} \gets \exp(\boldsymbol{\delta}_{l}^{\mathrm{in}})$
        \State $\mathbf{s}_{l}^{\mathrm{post}} \gets \exp(\boldsymbol{\delta}_{l}^{\mathrm{post}})$
    \EndFor

    \State Forward $\hat{\mathcal{M}}$ with the RMSNorm scaling vectors and its element-wise reciprocal
    \State Compute $\mathcal{L}$
    \State Update $\{\boldsymbol{\delta}_{l}^{\mathrm{in}}, \boldsymbol{\delta}_{l}^{\mathrm{post}}\}_{l=1}^{L}$

\EndFor

\For{$l = 1$ to $L$}
    \State Fuse $\mathbf{s}_{l}^{\mathrm{in}}$ and $\mathbf{s}_{l}^{\mathrm{post}}$ into the corresponding RMSNorm weights, and $(\mathbf{s}_{l}^{\mathrm{post}})^{-1}$ into the router weights
\EndFor

\State \Return $\hat{\mathcal{M}}$

\end{algorithmic}
\end{algorithm}

\section{Additional Implementation Details}
\label{app:implementations}
We randomly sample 128 sequences, each containing 2,048 tokens, from C4 using a fixed random seed of 0. The calibration set is used for expert-wise cost estimation prior to bit-width allocation, GPTQ calibration, and RoCo fine-tuning. We solve the ILP problem in Eq.~\ref{eq:ILP} using the HiGHS solver via SciPy. Following GEMQ, we apply GPTQ with a group size of 128 and asymmetric quantization under the final bit-width assignment. The clipping ranges are determined by minimizing the following weight quantization error:

\begin{equation}
\label{eq:clipping}
\min_{l,u}
\sum_{i,j}\left|W_{ij}-\hat{W}_{ij}(l,u)\right|^{2.4},
\end{equation}

where $l$ and $u$ denote the lower and upper clipping bounds, respectively, and $\hat{\mathbf{W}}(l,u)$ denotes the quantized-dequantized weights under the corresponding clipping range.

For Qwen3.5-35B-A3B, the maximum pruning ratio $\rho$ is set to 0.1, 0.15, and 0.3 for average expert bit-widths of 2.5, 2.0, and 1.5 bits, respectively, with a RoCo learning rate of $3\times10^{-3}$. In contrast, Qwen3-30B-A3B-Instruct-2507 is less sensitive to the maximum pruning ratio, maintaining strong performance even at $\rho=1.0$, which imposes no additional upper bound on the pruning ratio. We therefore set $\rho=1.0$ and use a RoCo learning rate of $5\times10^{-3}$ for this model. For both models, RoCo is optimized using AdamW for one epoch with a batch size of 1, $\beta_1=0.9$, $\beta_2=0.999$, and $\epsilon=10^{-8}$.

For downstream evaluation, we use EvalScope and follow the official sampling recommendations for both Qwen3-30B-A3B-Instruct-2507 and Qwen3.5-35B-A3B in non-thinking mode: temperature $=0.7$, top-$p=0.8$, top-$k=20$, and min-$p=0.0$. The maximum generation lengths are set to 8,192 tokens for GPQA-Diamond and MATH-500, 4,096 tokens for IFEval and GSM8K, and 2,048 tokens for MMLU-Redux and BFCL-v3.

\section{Efficiency Analysis}

\begin{figure}[H]
    \centering
    \includegraphics[width=\linewidth]{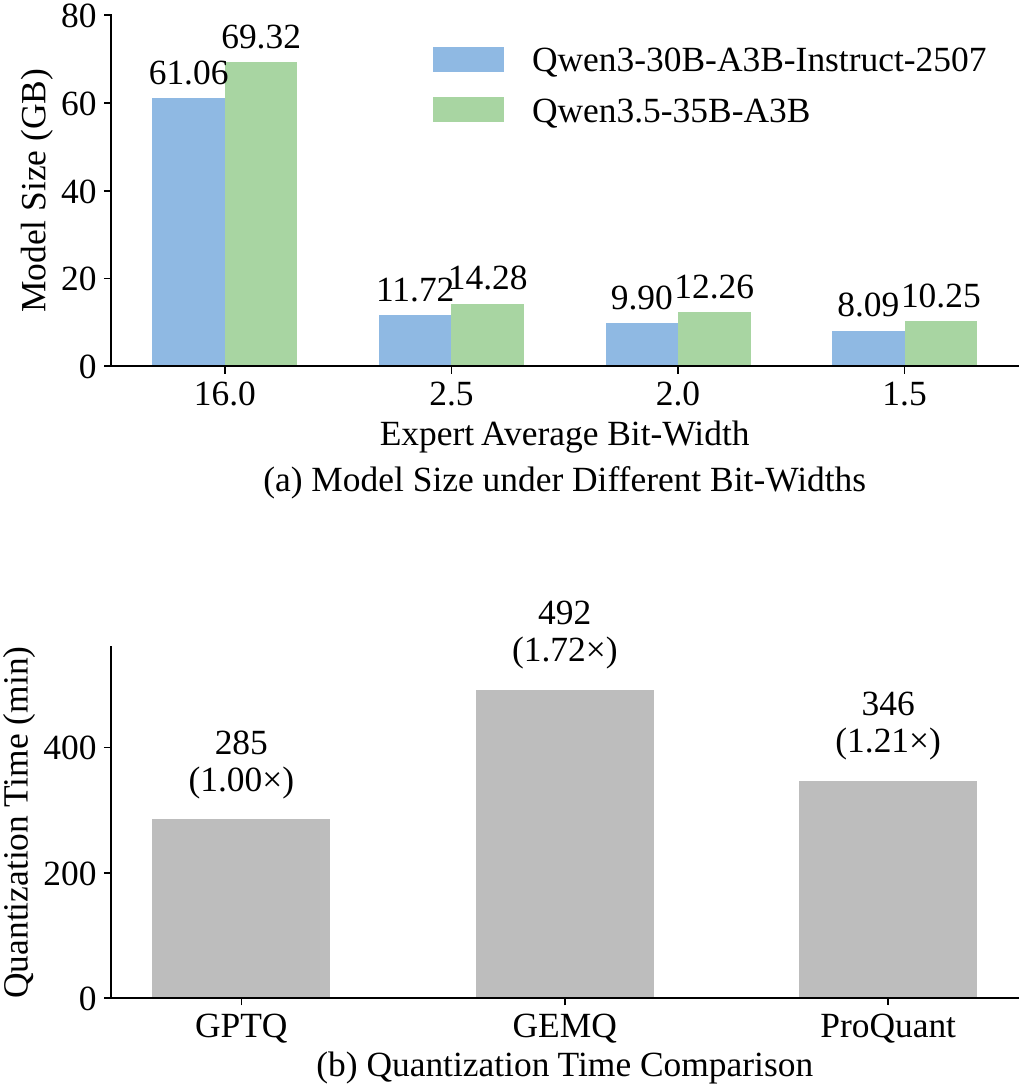}
    \caption{(a) Model size comparison of Qwen3-30B-A3B-Instruct-2507 and Qwen3.5-35B-A3B under different expert average bit-width settings. (b) End-to-end quantization time comparison of different methods on Qwen3-30B-A3B-Instruct-2507 under a 2.0-bit expert average bit-width setting.}
    \label{fig:model_size_and_quant_time}
\end{figure}

\textbf{Model Size.} As shown in Fig.~\ref{fig:model_size_and_quant_time}(a), ProQuant achieves substantial model compression under different expert average bit-width settings. Since the expert average bit-width does not directly reflect the overall storage cost due to non-expert components in MoE models, we report the model size after quantization. For Qwen3-30B-A3B-Instruct-2507, ProQuant reduces the BF16 model size from 61.06 GB to 8.09 GB, achieving 7.55× storage compression. Combined with the generation-based evaluation results, ProQuant effectively reduces storage requirements while maintaining strong downstream performance.

\textbf{Quantization Time.} We further compare the quantization time required to obtain the final quantized models. As shown in Fig.~\ref{fig:model_size_and_quant_time}(b), although ProQuant introduces additional statistical computation for bit-width allocation and RoCo fine-tuning compared with GPTQ, it only incurs a 1.21× overhead. Compared with GEMQ, which determines bit-width allocation through gradient-based backpropagation, ProQuant adopts a much lighter-weight allocation process. These results demonstrate that ProQuant improves low-bit model performance while maintaining high quantization efficiency.

\subsection{Model Size Calculation}

We calculate model size based on the parameter counts and quantization precision of different components. All original weights are stored in BF16, with 16 bits per parameter. For asymmetric quantization with a group size of 128, we account for two 16-bit metadata values per group, corresponding to the scale and zero-point. Accordingly, the storage ratio of $b$-bit quantized weights relative to their original BF16 representation is:
\begin{equation}
    r(b)=\frac{b}{16}+\frac{16+16}{128\times16}
=\frac{b}{16}+\frac{1}{64}.
\end{equation}

We divide model parameters into three categories according to their quantization configurations: experts subject to mixed-precision allocation, parameters fixed at 4-bit precision, and parameters retained in BF16. The total model size under an average expert bit-width of $B$ is calculated as:
\begin{equation}
S_{\mathrm{E}}r(B)
+
S_4r(4)
+
S_{\mathrm{FP}}, 
\end{equation}

where $S_{\mathrm{E}}$, $S_4$, and $S_{\mathrm{FP}}$ denote the original BF16 storage sizes of these three parameter categories, respectively, measured in GB. Here, $B$ denotes the parameter-weighted average expert bit-width. When expert pruning is considered, Eq. (15) provides a conservative upper bound on parameter storage, as it still accounts for quantization metadata associated with pruned experts.

\textbf{Qwen3-30B-A3B-Instruct-2507.}
The model contains 48 MoE layers with 128 routed experts per layer.
According to the quantization configuration, the expert weights, attention weights, and remaining BF16 parameters occupy $57.982$ GB, $1.812$ GB, and $1.270$ GB in their original BF16 representation, respectively.
Therefore, the model size under an average expert bit-width $B$ is

\begin{equation}
S_{\mathrm{Qwen3}}(B)
=
57.982\,r(B)
+
1.812\,r(4)
+
1.270 .
\end{equation}

The resulting model sizes are $11.717$ GB, $9.905$ GB, and $8.093$ GB under average expert bit-widths of $2.5$, $2.0$, and $1.5$ bits, respectively.

\textbf{Qwen3.5-35B-A3B.}
The model consists of 40 MoE layers with 256 routed experts and one shared expert per layer.
Its attention module combines 30 Gated DeltaNet layers and 10 full-attention layers.
Under our quantization configuration, the routed experts, shared experts, QKVO projections, and remaining BF16 parameters occupy $64.425$ GB, $0.252$ GB, $2.055$ GB, and $2.590$ GB in BF16 storage, respectively.
The corresponding model size is calculated as

\begin{equation}
\begin{aligned}
S_{\mathrm{Qwen3.5}}(B)
=&\,64.425\,r(B)\\
&+(0.252+2.055)\,r(4)\\
&+2.590 .
\end{aligned}
\end{equation}

This yields model sizes of $14.276$ GB, $12.263$ GB, and $10.249$ GB under average expert bit-widths of $2.5$, $2.0$, and $1.5$ bits, respectively.
The calculation considers only the language model parameters and excludes the vision encoder.

\section{Comparative Analysis of Expert Importance Metrics}

We compare the effects of different expert importance metrics on mixed-precision allocation and further examine the need to explicitly remove the influence of expert activation frequency. For a fair comparison, we use $\{1,2,3\}$ as the candidate expert bit-widths, fix the average expert bit-width at 2.0 bits, and keep the subsequent allocation procedure and all other quantization settings identical. No post-quantization finetuning is performed for any method.

As shown in Table~\ref{tab:allocation_ablation}, our importance metric achieves an average score of 61.40 across six benchmarks, outperforming the metrics used in MxMoE, PMQ, and GEMQ without finetuning by 11.51, 11.75, and 11.18 percentage points, respectively. These results demonstrate that the definition of expert importance substantially affects downstream performance under the same bit-width budget and allocation settings.

To examine the influence of activation frequency, we introduce PoP$^{*}$, which does not explicitly divide the expert quantization cost by the expert activation count, while keeping all other settings identical to PoP. Explicitly removing the influence of activation frequency increases the average score from 47.84 to 61.40 and improves performance on five of the six benchmarks, with a decrease on MMLU-Redux. We hypothesize that expert activation patterns on the C4 calibration set may be more closely aligned with those on MMLU-Redux. Such alignment could favor importance metrics that incorporate activation frequency on MMLU-Redux, while retaining this frequency dependence leads to lower performance on all other evaluated benchmarks.

\begin{table*}[t]
\centering
\small
\setlength{\tabcolsep}{3pt}

\begin{adjustbox}{max width=\textwidth}
\begin{tabular}{>{\centering\arraybackslash}p{1.6cm}|c|c|cccccc|c}
\hline
\textbf{Model}
& \textbf{Bit Budget}
& \textbf{Method}
& \textbf{GSM8K}
& \textbf{MATH-500}
& \textbf{GPQA-D}
& \textbf{IFEval}
& \textbf{MMLU-R}
& \textbf{BFCL-v3}
& \textbf{AVG} \\
\hline

\multirow{5}{*}{\parbox{1.6cm}{\centering Qwen3-30B-A3B-\\Instruct-2507}}
& \multirow{5}{*}{2.0}
& MxMoE
& 81.43 & 15.60 & 21.72 & 75.80 & \textbf{74.65} & 30.11 & 49.89 \\

& & PMQ
& 84.61 & 12.80 & 24.24 & 75.03 & 71.44 & 29.75 & 49.65 \\

& & GEMQ w/o FT
& 83.71 & 17.20 & \textbf{25.76} & 73.74 & 71.14 & 29.74 & 50.22 \\

& & PoP$^{*}$ ($\{1,2,3\}$)
& 86.13 & 15.40 & 21.72 & 73.22 & 61.98 & 28.61 & 47.84 \\

& & PoP ($\{1,2,3\}$)
& \textbf{91.74} & \textbf{78.60} & 22.22 & \textbf{77.87} & 58.18 & \textbf{39.76} & \textbf{61.40} \\

\hline
\end{tabular}
\end{adjustbox}

\caption{Comparison of expert importance metrics on Qwen3-30B-A3B-Instruct-2507 with an average expert bit-width of 2.0 bits. All methods use candidate bit-widths $\{1,2,3\}$ and the same allocation procedure, without post-quantization finetuning. PoP$^{*}$ omits explicit normalization by the expert activation count.}
\label{tab:allocation_ablation}
\end{table*}

\section{Dialog Examples}

\begingroup
\small
\newcommand{\generationexamplebreakable}[6]{%
  \begin{tcolorbox}[breakable,colback=white,colframe=white,boxrule=0pt,
    boxsep=0pt,left=8pt,right=8pt,top=0pt,bottom=0pt,
    before skip=8pt,after skip=10pt]
  \noindent\rule{\linewidth}{0.6pt}\par\smallskip
  #1\par
  \smallskip\noindent\rule{\linewidth}{0.25pt}\par\smallskip
  \noindent\textbf{BF16 response}\par #2\par
  \smallskip\noindent\rule{\linewidth}{0.25pt}\par\smallskip
  \noindent\textbf{MxMoE response}\par #3\par
  \smallskip\noindent\rule{\linewidth}{0.25pt}\par\smallskip
  \noindent\textbf{PMQ response}\par #4\par
  \smallskip\noindent\rule{\linewidth}{0.25pt}\par\smallskip
  \noindent\textbf{GEMQ response}\par #5\par
  \smallskip\noindent\rule{\linewidth}{0.25pt}\par\smallskip
  \noindent\textbf{ProQuant response}\par #6
  \par\smallskip\noindent\rule{\linewidth}{0.6pt}
  \end{tcolorbox}%
}

\generationexamplebreakable
{%
    \textbf{Example 1: MATH-500 (Level 2/5)}\par
    \textbf{Prompt:} A $90^\circ$ rotation around the origin in the counter-clockwise direction is applied to $7 + 2i.$ What is the resulting complex number?
    Please reason step by step, and put your final answer within \texttt{\textbackslash boxed\{\}}.
}
{We are given a complex number: $7+2i$, and we are to apply a $90^\circ$ counter-clockwise rotation around the origin.\par
 \textbf{Step 1: Understand the rotation in the complex plane.} In the complex plane, multiplying a complex number by $i$ corresponds to a $90^\circ$ counter-clockwise rotation about the origin. Why? Because $i=\cos(90^\circ)+i\sin(90^\circ)$; multiplying by $i$ rotates the number by $90^\circ$ counter-clockwise.\par
 \textbf{Step 2: Multiply the complex number by $i$.}
 $i(7+2i)=i\cdot7+i\cdot2i$;
 $=7i+2i^2$.
 Recall that $i^2=-1$, so $7i+2(-1)=7i-2=-2+7i$.\par
 \textbf{Final Answer:} \texttt{\textbackslash{}boxed\{-2 + 7i\}}. This is the result of rotating $7+2i$ by $90^\circ$ counter-clockwise around the origin.\par
 \smallskip\textcolor{green!50!black}{\textbf{Correct.}}}
{We are given a complex number: $7+2i$, and we are to apply a $90^\circ$ rotation counter-clockwise around the origin. Let’s break this down step by step.\par
 \textbf{Step 1: Understand what a $90^\circ$ counter-clockwise rotation means.} Rotating a point (or complex number) by $90^\circ$ counter-clockwise around the origin means we are transforming its coordinates by rotating it in the plane. In complex numbers, this is equivalent to rotating the vector from the origin (i.e., the complex number itself) by $90^\circ$.\par
 But here's the key: the rotation is around the origin, so we are rotating the vector $7+2i$ as a vector from the origin, by $90^\circ$ counter-clockwise.\par
 \textbf{Step 2: Recall how rotation affects complex numbers.} The real part is the $x$-axis. The imaginary part is the $y$-axis. So $7+2i$ is a point at $(7,2)$ in the plane.\par
 The vector $(7,2)$ is rotated so that its new coordinates become $(-2,7)$? Wait---let's think carefully. A $90^\circ$ CCW rotation of $(x,y)$ gives $(-y,x)$? Wait---let's verify. Take the vector $(1,0)$: rotating $90^\circ$ CCW gives $(0,1)$.\par
 \ldots\par
 \smallskip\textcolor{red!70!black}{\textbf{Incorrect: repeated rotation checks without a final answer.}}}
{We are given a complex number: $7+2i$, and we are to apply a $90^\circ$ rotation counter-clockwise around the origin to this number. Let's break this down step by step.\par
 \textbf{Step 1: Understand the meaning of $7+2i$.} The real part is $7$. The imaginary part is $2i$. So the complex number is $z=7+2i$. This is the same as $z=7+2j$ in engineering notation, but here $i$ is the imaginary unit.\par
 \textbf{Step 2: What does a $90^\circ$ counter-clockwise rotation mean?} Rotating a complex number by $90^\circ$ counter-clockwise around the origin is equivalent to multiplying the complex number by $i$. Wait---let's clarify this. Actually, in complex numbers, rotation is not done by multiplying by $i$. Instead, we use rotation in the plane, which is geometric.\par
 \textbf{Step 3: Apply rotation geometrically.} Rotation of a point $(x,y)$ by $90^\circ$ counter-clockwise: $(x,y)$ becomes $(y,x)$? Wait---let's recall the rotation matrix. The rotation matrix for 90° counter-clock clockwise is: \par
 \texttt{\textbackslash left \textbackslash left \textbackslash left}\ldots\par
 \smallskip\textcolor{red!70!black}{\textbf{Incorrect: malformed rotation-matrix loop; no final answer.}}}
{We are given a complex number: $7+2i$, and we are to apply a $90^\circ$ counter-clockwise rotation around the origin to it. We need to find the resulting complex number.\par
 \textbf{Step 1: Understand the complex number.} The complex number $7+2i$ means $7$ is the real part and $2i$ is the imaginary part. This is a point in the complex plane: $(7,2)$.\par
 \textbf{Step 2: Understand what a $90^\circ$ counter-clockwise rotation means.} We are rotating the point about the origin $(0,0)$, not about itself. A $90^\circ$ counter-clockwise rotation about the origin corresponds to multiplying the complex number by $i$. But wait: rotation about the origin is not the same as multiplying by $i$, unless you're rotating the vector from the origin. To rotate it around the origin, we must translate it to the origin? No---rotation about the origin means we keep the origin fixed.\par
 \textbf{Step 3: Use geometric rotation.} The new $x$-coordinate becomes $2$ (original $y$); the new $y$-coordinate becomes $7$ (original $x$), but negated because of rotation?\par
 \ldots\par
 I think the correct answer is $2+7i$. So, the resulting complex number is \texttt{\textbackslash{}boxed\{2 + 7i\}}.\par
 \ldots\par
 So, \texttt{\textbackslash{}boxed\{2 + 7i\}}.\par
 \ldots\par
 So, \texttt{\textbackslash{}boxed\{2 + 7i\}}.\par
 \ldots\par
 After $90^\circ$ CCW rotation, the new $x=y$, new $y=x$. So, the point at $(7,2)$ becomes $(2,7)$. Therefore, the resulting complex number is \texttt{\textbackslash{}boxed\{2 + 7i\}}.\par
 But I think it's wrong. Let's use the correct value. After $90^\circ$ CCW rotation, the new complex number is $2+7i$. So, \texttt{\textbackslash{}boxed\{2 + 7i\}}. I think that's the answer.\par
 \smallskip\textcolor{red!70!black}{\textbf{Incorrect: wrong sign after rotating $(7,2)$.}}}
{We are given a complex number: $7+2i$, and we are to apply a $90^\circ$ counter-clockwise rotation around the origin.\par
 \textbf{Step 1: Understand what a $90^\circ$ counter-clockwise rotation does to a complex number.} In the complex plane, rotating a complex number counter-clockwise by $90^\circ$ is equivalent to multiplying it by $i$. This is because $i=\cos(90^\circ)+i\sin(90^\circ)$, so multiplying by $i$ rotates the number by $90^\circ$ counter-clockwise.\par
 \textbf{Step 2: Multiply $7+2i$ by $i$.}
 $i(7+2i)=i\cdot7+i\cdot2i$;
 $=7i+2i^2$.
 Recall that $i^2=-1$, so $7i+2(-1)=7i-2$. So the result is $-2+7i$.\par
 \textbf{Final Answer:} \texttt{\textbackslash{}boxed\{-2 + 7i\}}.\par
 \smallskip\textcolor{green!50!black}{\textbf{Correct.}}}

\generationexamplebreakable
{%
    \textbf{Example 2: MATH-500 (Level 2/20)}\par
    \textbf{Prompt:} Simplify $\frac{\sec x}{\sin x}-\frac{\sin x}{\cos x}$.
    Please reason step by step, and put your final answer within \texttt{\textbackslash boxed\{\}}.
}
{We are asked to simplify the expression $\frac{\sec x}{\sin x}-\frac{\sin x}{\cos x}$.\par
 \textbf{Step 1: Recall trigonometric identities.} We know $\sec x=\frac{1}{\cos x}$.\par
 \textbf{Step 2: Rewrite $\sec x$ in terms of $\cos x$.} Replace $\sec x$ with $\frac{1}{\cos x}$. Simplify the first term:
 $\frac{\frac{1}{\cos x}}{\sin x}=\frac{1}{\cos x\sin x}$.
 The expression becomes $\frac{1}{\cos x\sin x}-\frac{\sin x}{\cos x}$.\par
 \textbf{Step 3: Get a common denominator.} Both terms have $\cos x$ in the denominator. The least common denominator is $\cos x\sin x$. Rewrite the second term:
 $\frac{\sin x}{\cos x}=\frac{\sin^2 x}{\cos x\sin x}$.\par
 \textbf{Step 4: Combine the fractions.}
 $\frac{1}{\cos x\sin x}-\frac{\sin^2 x}{\cos x\sin x}$;
 $=\frac{1-\sin^2 x}{\cos x\sin x}$.\par
 \textbf{Step 5: Use Pythagorean identity.} Since $1-\sin^2 x=\cos^2 x$, substitute to get $\frac{\cos^2 x}{\cos x\sin x}$.\par
 \textbf{Step 6: Simplify.} Cancel one $\cos x$ from numerator and denominator: $\frac{\cos x}{\sin x}=\cot x$.\par
 \textbf{Final Answer:} \texttt{\textbackslash{}boxed\{\textbackslash{}cot x\}}.\par
 \smallskip\textcolor{green!50!black}{\textbf{Correct.}}}
{We are given the expression $\frac{\sec x}{\sin x}-\frac{\sin x}{\cos x}$. Let’s simplify this step by step.\par
 \textbf{Step 1: Understand the terms.} Sec $x$ means the secant of angle $x$, which is $\frac{1}{\cos x}$. So $\frac{\sec x}{\sin x}$ is the same as $\frac{1/\cos x}{\sin x}$, which is $\frac{1}{\cos x\sin x}$. But let’s rewrite it clearly.\par
 Actually, sec $x$ is $\frac{1}{\cos x}$, so $\frac{\sec x}{\sin x}=\frac{1}{\sin x\cos x}$.\par
 Wait---let’s be careful. The expression is:
 \texttt{\textbackslash frac\{\textbackslash sec x\}\{\textbackslash sin x\} =}
 \texttt{\textbackslash frac \textbackslash text\{sec x\} / \textbackslash sin x}.\par
 But actually, that’s not quite right. Let’s clarify.\par
 \ldots\par
 But this is messy.\par
 \ldots\par
 \smallskip\textcolor{red!70!black}{\textbf{Incorrect: loops on fraction rewriting; no final answer.}}}
{We are given the expression $\frac{\sec x}{\sin x}-\frac{\sin x}{\cos x}$. But this notation is ambiguous and likely contains a misunderstanding. Let's carefully analyze what's going on.\par
 \textbf{Step 1: Understand the notation.} ``Sec $x$'' likely refers to the secant of $x$; ``sin $x$'' is sine and ``cos $x$'' is cosine. But this is not standard mathematical notation. It's written as if we are dividing ``sec $x$'' by ``sin $x$'', and subtracting ``sin $x$'' divided by ``cos $x$''.\par
 Wait---perhaps this is not a mathematical expression but a typographical or notational error?\par
 \textbf{Step 2: Consider trigonometric identities.} But perhaps the user meant $\sec x/\sin x$ as a ratio? But still, the expression is not standard. Could this be a miswritten version of a trigonometric identity or formula?\par
 \ldots\par
 The expression is not valid or cannot be simplified due to ambiguous and non-standard notation. Therefore, the expression cannot be simplified as written.\par
 \textbf{Final Answer:} \texttt{\textbackslash boxed\{Cannot be simplified\}}.\par
 \smallskip\textcolor{red!70!black}{\textbf{Incorrect: treats a valid expression as ambiguous.}}}
{We are given the expression $\frac{\sec x}{\sin x}-\frac{\sin x}{\cos x}$. Let’s simplify this step by step.\par
 \textbf{Step 1: Understand the terms.} We are dealing with trigonometric functions: $\sec x$ is the secant, $\sin x$ is the sine, and $\cos x$ is the cosine. But here, we are dealing with ratios of these values.\par
 \textbf{Step 2: Rewrite the expression.} Wait---there's a typo in the expression. But let's clarify: the expression is $\frac{\sec x}{\sin x}-\frac{\sin x}{\cos x}$. We are to simplify this expression as written.\par
 $\frac{\sec x}{\sin x}=\frac{1/\cos x}{\sin x}$;
 $=\frac{1}{\cos x\sin x}$. Wait---actually, this is ambiguous. $\sec x$ is a function, not a number.\par
 \ldots\par
 It doesn't simplify. Therefore, the simplified form is the expression itself. So the answer is:\par
 \texttt{\textbackslash{}boxed\{\textbackslash{}frac\{\textbackslash{}sec x\}\{\textbackslash{}sin x\}\allowbreak{} -\allowbreak{} \textbackslash{}frac\{\textbackslash{}sin x\}\{\textbackslash{}cos\}\}}.\par
 But that's not simplified.\par
 \ldots\par
 Final answer:\par
 \texttt{\textbackslash{}boxed\{\textbackslash{}frac\{\textbackslash{}sec x\}\{\textbackslash{}sin x\}\allowbreak{} -\allowbreak{} \textbackslash{}frac\{\textbackslash{}\ \textbackslash{}sin x\}\{\textbackslash{}cos\}\}}.\par
 But that's not simplified.\par
 \ldots\par
 \smallskip\textcolor{red!70!black}{\textbf{Incorrect: stops at an unsimplified expression.}}}
{We are asked to simplify the expression $\frac{\sec x}{\sin x}-\frac{\sin x}{\cos x}$.\par
 \textbf{Step 1: Recall identities.} We know $\sec x=\frac{1}{\cos x}$. So, rewrite $\sec x$ in terms of $\cos x$:
 $\frac{\sec x}{\sin x}=\frac{1/\cos x}{\sin x}$;
 $=\frac{1}{\cos x\sin x}$.
 Now the expression becomes $\frac{1}{\cos x\sin x}-\frac{\sin x}{\cos x}$.\par
 \textbf{Step 2: Find a common denominator.} The least common denominator is $\cos x\sin x$. Rewrite the second term:
 $\frac{\sin x}{\cos x}=\frac{\sin^2 x}{\cos x\sin x}$.
 Now subtract:
 $\frac{1}{\cos x\sin x}-\frac{\sin^2 x}{\cos x\sin x}$;
 $=\frac{1-\sin^2 x}{\cos x\sin x}$.\par
 \textbf{Step 3: Use identity $1-\sin^2 x=\cos^2 x$.}
 $\frac{1-\sin^2 x}{\cos x\sin x}=\frac{\cos^2 x}{\cos x\sin x}$.
 Now simplify:
 $\frac{\cos^2 x}{\cos x\sin x}=\frac{\cos x}{\sin x}=\cot x$.\par
 \textbf{Final Answer:} \texttt{\textbackslash{}boxed\{\textbackslash{}cot x\}}.\par
 \smallskip\textcolor{green!50!black}{\textbf{Correct.}}}

\generationexamplebreakable
{%
    \textbf{Example 3: BFCL-v3 (simple/1)}\par
    \textbf{Prompt:} Calculate the factorial of 5 using math functions.\par
    \textbf{Tool:} \texttt{math\_factorial(number: integer)}. Description: Calculate the factorial of a given number (Python 3 syntax). The required \texttt{number} argument is an integer: the number for which factorial needs to be calculated.
}
{\texttt{[\{"function":\{}\par
 \texttt{"arguments":\{"number":5\},}\par
 \texttt{"name":"math\_factorial"\},}\par
 \texttt{"id":"chatcmpl-tool-\allowbreak{}1e13c937\allowbreak{}663d4ebe\allowbreak{}b0ae5f34\allowbreak{}96935db8",}\par
 \texttt{"internal":null,"parse\_error":null,}\par
 \texttt{"type":null,"view":null\}]}\par
 \smallskip\textcolor{green!50!black}{\textbf{Correct.}}}
{\texttt{\textless tool\_call\textgreater}\par
 \texttt{\{"name": math\_factorial,}\par
 \texttt{"arguments": math\_factorial, 5)}\par
 \texttt{\textless/tool\_call\textgreater}\par
 \smallskip\textcolor{red!70!black}{\textbf{Incorrect: malformed tool-call arguments.}}}
{\texttt{\textless tool\_call\textgreater}\par
 \texttt{\{"name": "mathdba",}\par
 \texttt{"arguments": \{"number": \{"type": "integer",}\par
 \texttt{"description": "The number for which factorial needs to be calculated."\%\}\}}\par
 \texttt{\textless/tool\_call\textgreater}\par
 \smallskip\textcolor{red!70!black}{\textbf{Incorrect: wrong function name and malformed arguments.}}}
{\texttt{[\{"function":\{}\par
 \texttt{"arguments":\{"number":"number"\},}\par
 \texttt{"name":"math\_factorial"\},}\par
 \texttt{"id":"chatcmpl-tool-\allowbreak{}d32f30a0\allowbreak{}2e1d4fd8\allowbreak{}a066626a\allowbreak{}e647681b",}\par
 \texttt{"internal":null,"parse\_error":null,}\par
 \texttt{"type":null,"view":null\}]}\par
 \smallskip\textcolor{red!70!black}{\textbf{Incorrect: passes a string instead of the integer 5.}}}
{\texttt{[\{"function":\{}\par
 \texttt{"arguments":\{"number":5\},}\par
 \texttt{"name":"math\_factorial"\},}\par
 \texttt{"id":"chatcmpl-tool-\allowbreak{}2b59039d\allowbreak{}87e24cac\allowbreak{}98c20e5c\allowbreak{}f00a4494",}\par
 \texttt{"internal":null,"parse\_error":null,}\par
 \texttt{"type":null,"view":null\}]}\par
 \smallskip\textcolor{green!50!black}{\textbf{Correct.}}}

\generationexamplebreakable
{%
    \textbf{Example 4: BFCL-v3 (simple/6)}\par
    \textbf{Prompt:} What are the roots of the quadratic equation where $a=2$, $b=5$ and $c=3$?\par
    \textbf{Tool:} \texttt{solve\_quadratic(a: integer,\allowbreak{} b: integer,\allowbreak{} c: integer)}. Description: Find the roots of a quadratic equation and return both roots (Python 3 syntax). Required integer arguments: \texttt{a}, coefficient of $x^2$; \texttt{b}, coefficient of $x$; and \texttt{c}, constant term.
}
{\texttt{[\{"function":\{}\par
 \texttt{"arguments":\{"a":2,"b":5,"c":3\},}\par
 \texttt{"name":"solve\_quadratic"\},}\par
 \texttt{"id":"chatcmpl-tool-\allowbreak{}4157e7ac\allowbreak{}a65c477a\allowbreak{}bd6658c2\allowbreak{}09fb7164",}\par
 \texttt{"internal":null,"parse\_error":null,}\par
 \texttt{"type":null,"view":null\}]}\par
 \smallskip\textcolor{green!50!black}{\textbf{Correct.}}}
{Do you want to solve the quadratic equation with $a=2$, $b=5$, and $c=3$?\par
 The quadratic equation is $2x^2+5x+3=0$. We can solve this equation using the quadratic formula:\par
 \texttt{x = \textbackslash frac\{-b \textbackslash pm}\par
 \texttt{\textbackslash sqrt\{b\textasciicircum 2 - c\textasciicircum 2) \} / (2a)}\par
 Let's solve it step-by-step. First, compute the discriminant:
 $b^2-c^2=5^2-3^2=25-9=16$.\par
 Then compute the roots:
 \texttt{x = \textbackslash frac(-5 \textbackslash pm \textbackslash sqrt(16)) / (2*2)}\par
 \texttt{= (-5 \textbackslash pm 4) / 4}.\par
 $x=(-5+4)/4=-1/4$; $x=(-5-4)/4=-9/4$.\par
 So the roots are $x=-0.25$ and $x=-2.25$. Therefore, the roots are $-0.25$ and $-2.25$.\par
 \smallskip\textcolor{red!70!black}{\textbf{Incorrect: no tool call; incorrect roots.}}}
{\texttt{\textless tool\_call\textgreater}\par
 \texttt{\{"name": "solve\_quadratic",}\par
 \texttt{"arguments": \{"a": 2, "b": 5, "c": 3"\}}\par
 \texttt{\textless/tool\_call\textgreater}\par
 \smallskip\textcolor{red!70!black}{\textbf{Incorrect: malformed tool-call JSON.}}}
{\texttt{\textless tool\_call\textgreater}\par
 \texttt{\{"name": "solve\_quadratic",}\par
 \texttt{"arguments": \{"a": and "b" and "c": 2, 5, 3\}\}}\par
 \texttt{\textless/tool\_call\textgreater}\par
 \smallskip\textcolor{red!70!black}{\textbf{Incorrect: malformed argument mapping.}}}
{\texttt{[\{"function":\{}\par
 \texttt{"arguments":\{"a":2,"b":5,"c":3\},}\par
 \texttt{"name":"solve\_quadratic"\},}\par
 \texttt{"id":"chatcmpl-tool-\allowbreak{}e455b126\allowbreak{}f3664211\allowbreak{}b67c43e5\allowbreak{}2594e61f",}\par
 \texttt{"internal":null,"parse\_error":null,}\par
 \texttt{"type":null,"view":null\}]}\par
 \smallskip\textcolor{green!50!black}{\textbf{Correct.}}}

\captionof{table}{Qualitative comparison of model responses on four examples using Qwen3-30B-A3B-Instruct-2507. All quantized methods have an average expert bit-width of 2.5 bits and use the same prompt and decoding settings within each example. For readability, responses have been abridged and reformatted, with repetitive passages and off-topic digressions omitted while retaining key reasoning steps. For the two BFCL examples, tool definitions are summarized rather than reproduced in full JSON format.}\label{tab:generation_examples}
\endgroup

\section{License and Attribution}

Our implementation builds on GEMQ~\citep{deng2026gemq}, which is released under the MIT License. We retain the original copyright and license notices in our released code.

\section{AI Use Statement}

AI assistants were used only to modify code according to our instructions and to polish the manuscript wording. The literature review and development of the research contributions were conducted by the authors.

\end{document}